\documentclass[10pt,twocolumn,letterpaper]{article}

\usepackage[pagenumbers]{wacv} 

\definecolor{wacvblue}{rgb}{0.21,0.49,0.74}
\usepackage[pagebackref,breaklinks,colorlinks,allcolors=wacvblue]{hyperref}

\usepackage{float}

\def\wacvPaperID{672} 
\def\confName{WACV}
\def\confYear{2027}

\title{Bridging the Catalog-to-Real Gap: Scalable Product Recognition via Multi-Stage Contrastive Learning}

\author{Anyi Zhang\\
Walmart Global Tech\\
Sunnyvale, CA, USA\\
{\tt\small anyi.zhang@walmart.com}
\and
Joy Mazumder\\
Walmart Global Tech\\
Sunnyvale, CA, USA\\
{\tt\small joy.mazumder@walmart.com}
\and
Kiril Lomakin\\
Walmart Global Tech\\
Sunnyvale, CA, USA\\
{\tt\small kiril.lomakin@walmart.com}
}

\begin{document}
\maketitle
\begin{abstract}
Automated product recognition is a cornerstone of modern retail intelligence; however, accurately matching real-world, in-store images against extensive corporate catalogs remains a major scalability bottleneck for large-scale applications. In this work, we address this challenge by reformulating the task as an embedding-based cross-domain retrieval problem rather than a standard closed-set classification task. Specifically, we define the objective as retrieving the most corresponding catalog reference image for a given real-world product query crop from an expansive inventory. To bridge the severe domain gap between pristine studio packshots and noisy in-store queries, we introduce a novel catalog-to-real multi-stage contrastive learning paradigm (\textbf{Cat2Real}). This framework fine-tunes a vision backbone by systematically exploiting both item-level and image-level similarities to drive targeted hard negative mining. Extensive empirical evaluations demonstrate that our paradigm scales seamlessly to unseen products and categories, yielding outstanding zero-shot generalization performance even in the complete absence of real-world training images for novel inventory. The fine-tuned model is made available under Apache 2.0 license\footnote{https://huggingface.co/zhanganyi88/Cat2Real-DINOv3-384}.
\end{abstract}
    
\section{Introduction}
\label{sec:intro}

\begin{figure*}[t]
  \centering
    \includegraphics[width=0.9\linewidth]{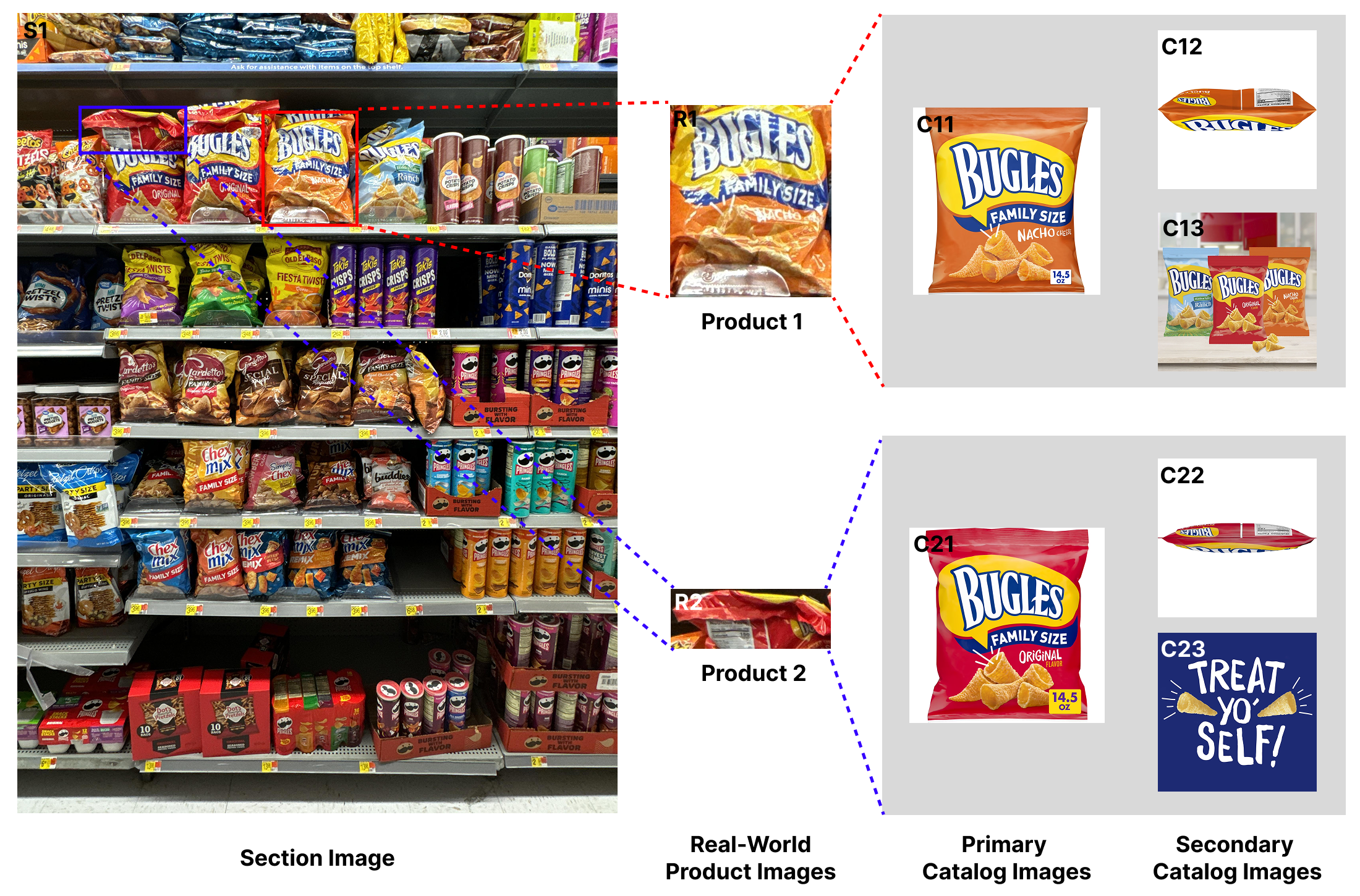}
   \caption{\textbf{Overview of the embedding-based product recognition paradigm for scalable retail deployment.} The task is formulated as matching noisy, real-world product crops (left) against an existing catalog reference database containing primary studio packshots and alternative secondary images (right) within a shared latent space. This formulation bridges the severe domain gap without requiring manual annotations for new inventory.}
   \label{fig:1}
\end{figure*}

Automated product recognition is a core catalyst for modern retail intelligence, enabling optimized supply chains and seamless customer experiences. While the initial pipeline step of localizing products is largely solved by modern object detectors, scaling the subsequent classification phase to global enterprise inventories remains a formidable bottleneck. This challenge stems from three compounding factors: \textbf{(1) Extreme Scale and Volatility:} Hypermarkets stock millions of active Stock Keeping Units (SKUs) that undergo daily inventory additions and frequent packaging redesigns, rendering static closed-set classifiers obsolete. \textbf{(2) Fine-Grained Ambiguity:} Pervasive inter-class similarity exists across retail categories, where distinct items share near-identical profiles differing only by subtle visual cues, such as flavor, size, or minor textual details. \textbf{(3) Environmental Noise:} Real-world in-store images suffer from low resolution, severe occlusion, arbitrary rotations, motion blur, and glare from protective glass displays.

To circumvent these challenges, a viable production system must operate under two core constraints: it must rely exclusively on pristine, pre-existing catalog packshots to eliminate the prohibitive costs of manual in-store annotation; and it must absorb real-time inventory updates without requiring model retraining. Consequently, we reformulate product recognition from a traditional closed-set classification problem into an open-set, cross-domain retrieval task. The objective is to project both noisy real-world product crops and high-quality catalog reference images into a shared latent embedding space where matching is performed via metric distance (see \cref{fig:1}). 

Because off-the-shelf foundation vision models lack sufficient discriminatory power across this severe domain gap, we introduce a novel multi-stage Catalog-to-Real (\textbf{Cat2Real}) contrastive learning framework. Unlike standard contrastive paradigms that rely on random negative sampling, which often yields uninformative or contradictory gradients in fine-grained tasks, Cat2Real systematically optimizes a vision backbone across three progressive stages. It utilizes item- and image-level similarities alongside dynamic nearest-neighbor selection as core mechanisms to filter corrupted queries, drive targeted hard negative mining, and match optimal positive views to eliminate training noise. Optimized via a triplet margin loss, this approach ensures the encoder learns robust, invariant semantic boundaries. Empirical evaluations demonstrate that Cat2Real significantly outperforms existing commercial and open-source embedding baselines (\cref{tab:1}). Crucially, our framework exhibits robust zero-shot generalization, scaling seamlessly to entirely unseen categories and products with minimal performance decay, making it exceptionally suited for continuous, large-scale retail deployment.

\section{Related Work}
\label{sec:related_work}

\begin{table*}[t]
  \centering 
        \begin{tabular}{@{}lcccr@{}}
          \toprule
          Method & Arch. &  Image Size & Top-1 Acc. & Top-5 Acc. \\
          \midrule
          Gemini-Embedding-2~\cite{geminiembedding2} & N/A & N/A & 68.71 & 89.34\\
          Multimodalembedding@001 & N/A & 512 & 63.51 & 84.39\\
          DINOv2~\cite{dinov2} & ViT-L/14 & 224 & 35.53 & 61.39\\
          DINOv3~\cite{dinov3} & ViT-L/16 & 224 & 49.77 & 77.18\\
          DINOv3~\cite{dinov3} & ViT-L/16 & 384 & 53.83 & 78.12\\
          \hline
          Cat2Real-DINOv2 & ViT-L/14 & 224 & 76.34 & 93.13\\
          Cat2Real-DINOv3-224 & ViT-L/16 & 224 & 78.80 & 94.41\\
          Cat2Real-DINOv3-384 & ViT-L/16 & 384 & {\bfseries 80.73} & {\bfseries 94.70}\\
          \bottomrule
        \end{tabular}
        \caption{\textbf{Main comparative evaluation results on the in-house Product Recognition Evaluation benchmark.} Evaluation metrics compare open-source foundation backbones (DINOv2, DINOv3) and top-tier commercial embedding models against our fine-tuned Cat2Real models across Top-1 and Top-5 accuracy.}
        \label{tab:1}
\end{table*}

\subsection{Product Recognition in Retail Industry}
Modern retail product recognition typically employs a two-step pipeline: object localization followed by fine-grained classification. While localization has become robust through models like YOLOv5~\cite{FAM1,khanam2024yolov5deeplookinternal}, YOLOv8~\cite{10528273,10533619}, and Faster R-CNN~\cite{NIPS2015_14bfa6bb,sinha2022improveddeeplearningapproach}, classification remains challenging due to extreme visual similarity and the need to scale across millions of products. Recent research has shifted toward embedding-based methods using triplet loss~\cite{sinha2022improveddeeplearningapproach} or multi-modal approaches that fuse visual features with OCR-extracted text~\cite{MVA1,OUCHEIKH2022115942}. Despite these advances, practical deployment is often hindered by the scarcity of annotated real-world data and environmental factors like occlusion or poor lighting.

\subsection{Image Encoders}
The selection of a high-performing foundational image encoder is critical for effective transfer learning, particularly in specialized and fine-grained domains. The transition from localized convolutional architectures~\cite{cnn,resnet} to global Vision Transformers (ViTs)~\cite{vit} has been fundamentally driven by massive-scale self-supervision. While language-supervised models like CLIP~\cite{clip} excel at high-level semantic alignment through image-text pairs, self-supervised frameworks such as Masked Autoencoders (MAE)~\cite{mae} prioritize detailed representation through pixel reconstruction. A central advancement in this field is the DINO series,~\cite{dino,dinov2,dinov3} which employs a self-distillation paradigm. DINOv2~\cite{dinov2} and DINOv3~\cite{dinov3} have emerged as powerful foundation models, utilizing patch-level objectives and Gram anchoring to capture both global context and fine-grained local structures, facilitating the creation of robust modality-agnostic feature spaces.

\subsection{Contrastive Learning}
Contrastive learning is particularly well-suited for few-shot classification problems~\cite{metalearning} like product recognition, where labeled data is scarce and classes frequently change. This paradigm in the visual domain aims to learn robust representations by pulling semantically similar samples together while pushing negatives apart in a latent space. Frameworks like SimCLR~\cite{SimCLR} and MoCo~\cite{moco} leverage NT-Xent~\cite{nt_xent} or InfoNCE~\cite{infonce} losses to contrast augmented views of an image against in-batch negatives. Similarly, triplet loss~\cite{facenet, triplet} learns discriminative features by enforcing a predefined margin between anchor, positive, and negative embeddings. Ultimately, these methodologies allow image encoders to map disparate data distributions into a unified space optimized for cross-domain matching and retrieval.
\section{Proposed Approach}
\label{sec:proposed_approach}

\begin{figure*}[t]
  \centering
    \includegraphics[width=0.9\linewidth]{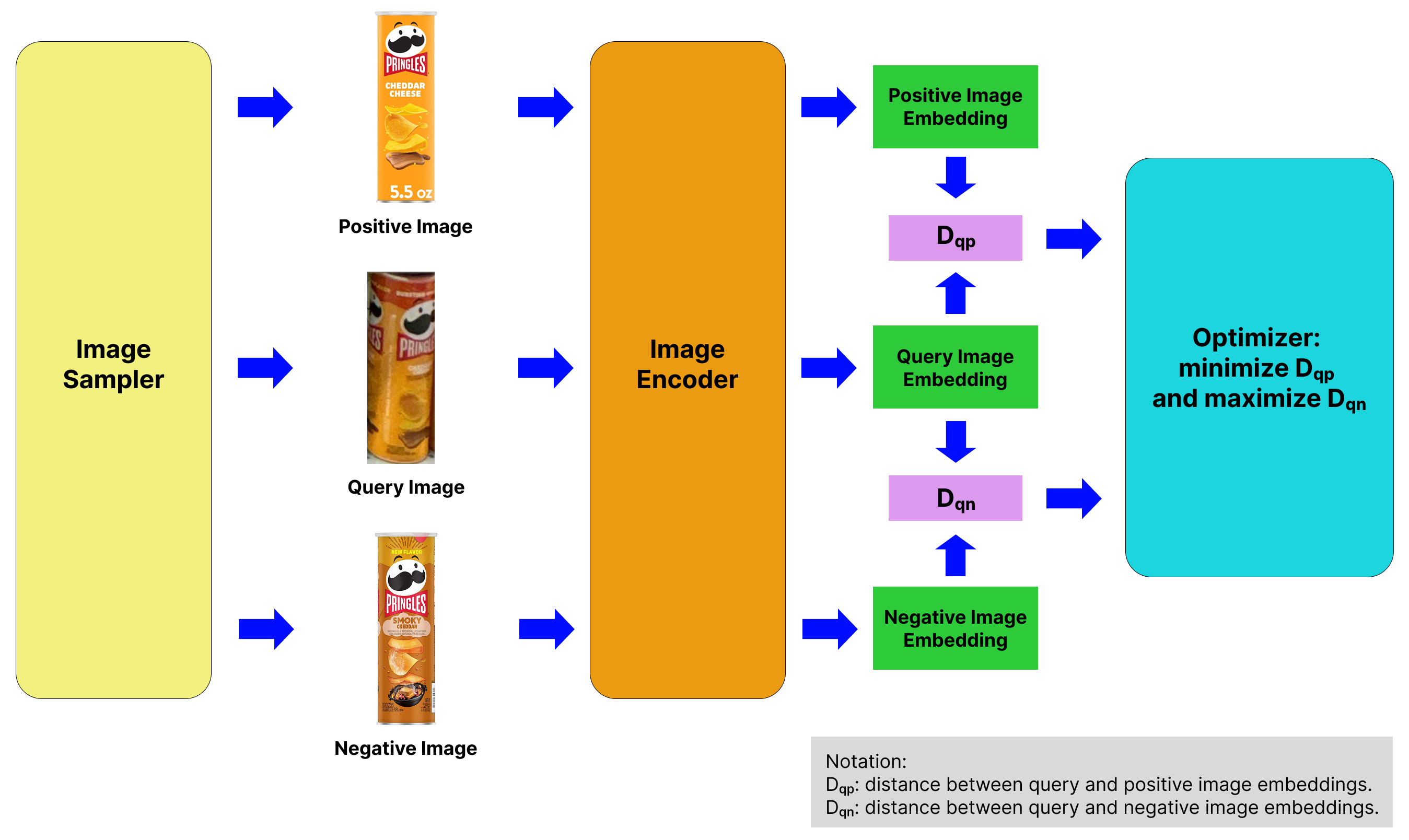}
    \caption{
      \textbf{Schematic representation of the contrastive learning training pipeline.} Real-world query images are paired with strategic positive and negative catalog samples via an automated image sampler. The encoder is optimized using a triplet margin loss that minimizes the distance ($D_{\text{qp}}$) between the query and the positive embedding while maximizing the distance ($D_{\text{qn}}$) to the negative embedding.
    }
    \label{fig:2}
\end{figure*}

\subsection{Problem Statement}

We aim to recognize retail products at scale by matching real-world images to a reference catalog. As illustrated in \cref{fig:1}, the pipeline begins with capturing full-section images for tasks such as planogram compliance. We employ object detection models to extract bounding boxes of individual products. Because these sections are captured in a single view, the resulting real-world crops are often low-resolution, blurry, rotated, or partially occluded, frequently rendering textual details unreadable.

To ensure system scalability and eliminate the need for manual labeling of new inventory, our objective is to map these real-world crops to specific product IDs using only existing catalog data. The catalog contains two categories of images: \textit{primary} (standardized front-facing packshots) and \textit{secondary} (alternative views, advertising materials, or irrelevant images). 

This task presents a significant domain gap challenge: real-world images suffer from environmental noise and quality degradation, while catalog packshots are high-resolution and studio-quality. Furthermore, the catalog itself may contain noise, such as missing packshots, limited viewing angles, or mislabeled images where the provided photo does not match the product ID (e.g., image \textit{C13} in \cref{fig:1}). Our goal is to develop an embedding-based matching framework robust to these distribution shifts and data inconsistencies.

\subsection{Training Paradigm}
\label{sec:training_paradigm}

To address the challenges of large-scale product recognition, we train an image encoder to map both real-world and catalog images into a shared embedding space. At inference, the encoder generates an embedding for a real-world product image, which is then used to identify the product ID via a similarity search against a vector database of catalog embeddings. Formally, the inference process is defined as:

\begin{equation}
    o_i^* = \operatorname{argmin}_{x \in X} \text{Dist}[Enc(r_i), Enc(c_j^x)]
\end{equation}

where \( o_i^*\) is the predicted product ID for the real-world image \(r_i\), \(X\) is the set of candidate products, \(c_j^x\) represents the \(j\)-th catalog image of product \(x\), \(Enc\) is the image encoder, and \(\text{Dist}\) is the distance metric.

The encoder is optimized using a contrastive learning paradigm (see \cref{fig:2}) designed to pull embeddings of the same product together while pushing those of different products apart. To bridge the significant domain gap between real-world and catalog distributions, we utilize a labeled dataset of real-world crops paired with their corresponding catalog packshots. During training, real-world images serve as queries, while catalog images provide the positive and negative images.

Unlike conventional contrastive methods that rely on random sampling, our approach addresses the noise inherent in catalog data and the necessity of fine-grained discrimination. Standard random sampling often fails when catalog images differ significantly from real-world query views or when products share nearly identical visual features. We therefore implement strategic data sampling techniques (discussed in \cref{sec:sampling}) to intentionally select suitable positives and hard negatives. The model is optimized using the triplet margin loss:

\begin{equation}
\begin{split}
\mathcal{L} = \sum_{i=1}^{N} \max \Big( 0, & \|Enc(q_i) - Enc(p_i)\|_2^2 \\
&- \|Enc(q_i) - Enc(n_i)\|_2^2 + \alpha \Big)
\end{split}
\end{equation}

where \( q_i\), \( p_i\), and \( n_i\) denote the query, positive, and negative embeddings, respectively, and \( \alpha\) is the margin hyperparameter.

\subsection{Data Sampling Techniques}
\label{sec:sampling}

\begin{figure*}[t]
  \centering
    \includegraphics[width=0.9\linewidth]{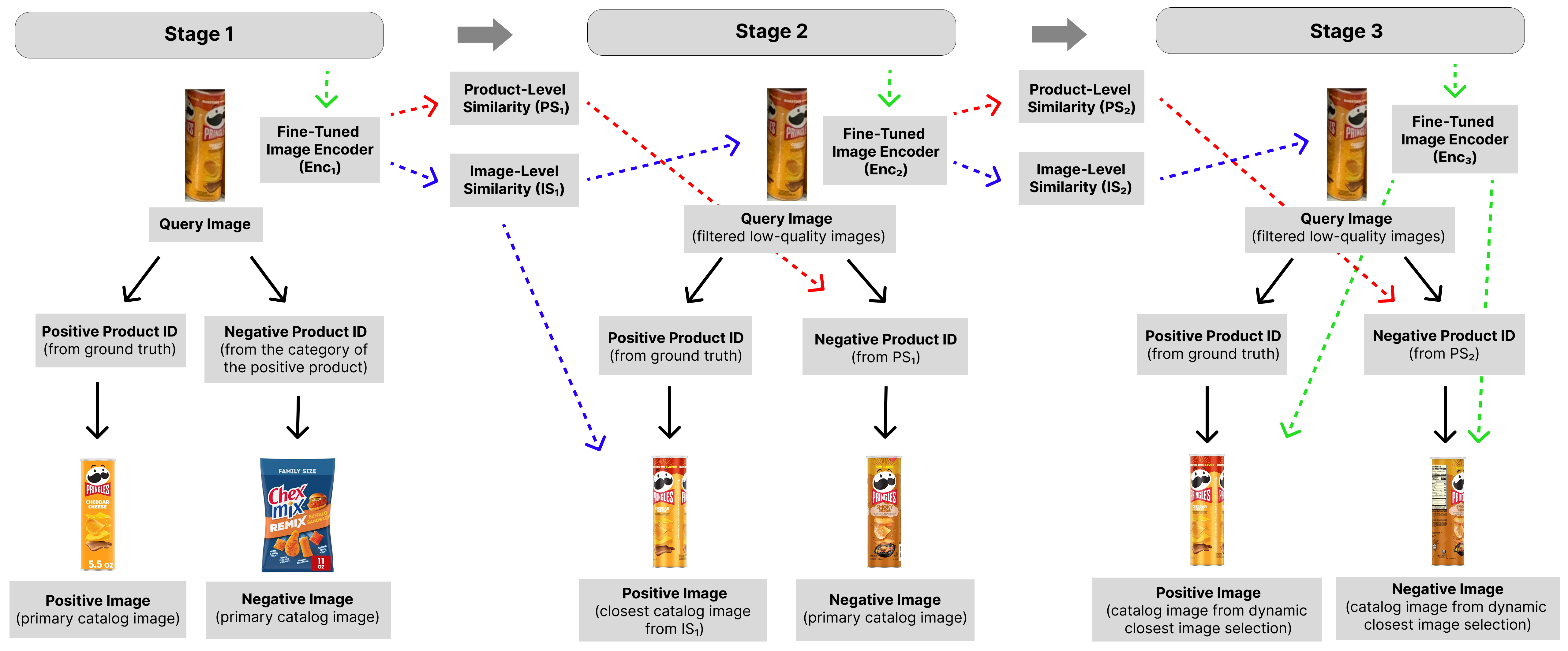}
    \caption{
      \textbf{The proposed three-stage iterative fine-tuning framework.} Stage 1 establishes baseline sampling using primary catalog pairs. Stage 2 introduces hard negative mining guided by static image-level ($\text{IS}$) and product-level ($\text{PS}$) similarity indices. Stage 3 incorporates a dynamic, on-the-fly catalog image selection strategy to continuously refine the feature space.
    }
    \label{fig:3}
\end{figure*}

A single real-world query image can theoretically be paired with multiple positive and negative catalog images. However, to ensure training efficiency, we select only one positive and one negative sample for each query image per epoch. The strategy for selecting these pairs is fundamental to achieving high recognition performance. While the positive sample must belong to the ground-truth product, catalog datasets often contain non-representative or low-quality images that are unsuitable for training. Furthermore, because query images may capture arbitrary viewpoints, it is critical to select a catalog image with a matching perspective when available; failing to do so can lead to conflicting gradients that destabilize the optimization process.

Using Product 2 in ~\cref{fig:1} as an example, the real-world crop \textit{R2} and catalog image \textit{C22} are both side-views, making \textit{C22} the ideal positive sample. If \textit{C21} (a front-view) were selected as the positive sample while \textit{C12} (a similar side-view of a different product) were used as the negative sample, the model would simultaneously attempt to minimize $Dist[Enc(R2), Enc(C21)]$ and maximize $Dist[Enc(R2), Enc(C12)]$. This contradictory objective can make optimization highly unstable. Identifying truly hard negatives is also challenging, as most products in the catalog appear visually distinct from the query, and even within a single product ID, catalog images can vary significantly.

To address these challenges, we propose a three-stage iterative framework to refine the image encoder. Each stage employs a progressively more sophisticated sampling strategy:

\textbf{Stage 1: Baseline Sampling.} The positive sample is restricted to the primary catalog image of the ground-truth product. To select a negative sample, we first identify a product from the same category as the ground truth and then utilize its primary catalog image. After Stage 1, the fine-tuned encoder $Enc_1$ is used to generate two similarity indices: product-level similarity ($PS_1$), representing the similarity between primary product images, and image-level similarity ($IS_1$), representing the similarity between query images and their respective ground-truth catalog images.

\textbf{Stage 2: Hard Negative Mining.} We first filter the query images using $IS_1$. If a query's highest image-level similarity falls below a threshold $T_I$, we conclude that the image is either of insufficient quality or that a matching view is missing from the catalog; such images are discarded to reduce training noise. For the remaining queries, the positive sample is the catalog image with the highest similarity to the query according to $IS_1$. Negative products are sampled from the top $N$ similar products in $PS_1$, where the probability $P_i$ of selecting product $i$ is defined as:
\begin{equation}
P_i = \frac{N - i + 1}{\sum_{j=1}^{N} j}
\end{equation}
where $i$ is the similarity rank within the top $N$ products. This weighted sampling prioritizes visually similar products to force the model to learn fine-grained distinctions without sacrificing dataset diversity. The primary image of the chosen negative product serves as the negative sample. Following this stage, we obtain an updated encoder $Enc_2$ and revised similarity indices $PS_2$ and $IS_2$.

\textbf{Stage 3: Dynamic Image Selection.} This stage utilizes the same filtering and negative product sampling logic as Stage 2 but introduces a dynamic, computationally intensive image selection strategy. Rather than using a static mapping, we use the encoder's latest weights during each mini-batch to calculate similarities between real-world crops and catalog images on the fly. The positive sample is dynamically set as the closest catalog image from the ground-truth product, and the negative sample is the closest catalog image from the selected negative product. This real-time refinement minimizes view-based noise and maximizes the effectiveness of hard negatives.

\section{Experiments}
\label{sec:experiments}

\begin{table*}[t]
  \centering 
    \begin{tabular}{@{}lccccr@{}}
      \toprule
      Method & Total Cat. & Total Products &  Total Real Images & Top-1 Acc. & Top-5 Acc.\\
      \midrule
      {\bfseries Cat2Real-DINOv3-384} & {\bfseries 52} & {\bfseries 45,243} & {\bfseries 196,531} & {\bfseries 80.73} & {\bfseries 94.70}\\
      Remove Eval Products & 52 & 43,330 & 180,946 & 80.23 & 94.20\\
      Batch 4  & 31 & 23,096 & 122,043 & 78.89 & 93.97\\
      Batch 3 & 22 & 16,843 & 94,230 & 78.19 & 93.18\\
      Batch 2 & 12 & 13,865 & 77,450 & 76.87 & 93.06\\
      Batch 1 & 6 & 11,206 & 68,407 & 75.77 & 92.84\\
      \bottomrule
    \end{tabular}
    \caption{\textbf{Zero-shot generalization performance.} Accuracy trajectories across training configurations with incrementally scaled category and product volumes (Batches 1--4). The ''Remove Eval Products'' configuration explicitly excludes all evaluation targets from the training phase. This clearly highlights the model's robust zero-shot adaptability to entirely unseen products and categories.}
    \label{tab:scaling}
\end{table*}

\subsection{Data Preparation}
To construct the training and evaluation datasets, real-world product images must be mapped to their corresponding product identifiers (IDs). This data collection and annotation process is conducted via two primary approaches:
\begin{itemize}  
\item \textbf{Method 1 (Section-Based Annotation):} Full-section retail rack images are captured, from which individual products are manually annotated with bounding boxes and product IDs.
\item \textbf{Method 2 (Barcode-Assisted Capture):} A dedicated mobile application is utilized to simultaneously scan a product's barcode and capture its real-world image, automating label assignment.
\end{itemize}
The training dataset incorporates images acquired from both methods, whereas the evaluation dataset exclusively comprises real-world product images collected via Method 1 to reflect true in-store shelf deployment scenarios. Comprehensive statistics for both sets are provided in \cref{tab:dataset_stat} in the Supplementary Material. Specifically, the training set consists of 196,531 real-world images spanning 26,910 unique products, with negative samples drawn from a pool of 45,243 unique catalog products. The evaluation set contains 5,043 real-world images across 1,865 unique products, operating within a candidate search scope of 2,155 unique products.

\subsection{Implementation Details}
We fine-tune the pretrained foundation models on our large-scale retail dataset using the AdamW optimizer. Training is conducted on a single NVIDIA A100 GPU with a batch size of 16. To optimize throughput, we employ 8 data loader workers and accumulate gradients over 8 steps, resulting in an effective batch size of $s = 128$. The Vision Transformer (ViT-L) backbone comprises 24 transformer blocks; to preserve general feature representations and reduce computational overhead, the weights of the first 12 layers are frozen, and only the remaining 12 layers are updated during fine-tuning. 

Furthermore, we employ layer-wise learning rate decay with a decay factor of $r = 0.75$. The learning rate $lr_{i}$ for the $i$-th layer from the top is defined as:
\begin{equation}
lr_{i} = 10^{-5} \times \frac{s}{256} \times r^{i-1}
\end{equation}
A cosine annealing scheduler with a linear warm-up phase is used to modulate the learning rate. To stabilize training, gradient clipping is enforced with an L2 norm threshold of 1.0. Mixed-precision training (FP16) is utilized to optimize memory efficiency and accelerate convergence. 

To bridge the domain gap, extensive data augmentation is applied. In addition to real-world images, catalog images are also used as query images after augmenting via random rotation, cropping, blurring, and contrast adjustments to simulate real-world variations. Positive and negative pairs undergo identical augmentation pipelines. To balance dataset diversity and mining stringency, the hyperparameter $N$ for negative product candidates is set to 50, and the similarity threshold for filtering low-quality images is established at 0.85.

\subsection{Results}

\begin{table}[t]
  \centering 
    \begin{tabular}{@{}lcccr@{}}
      \toprule
      Method & Top-1 Acc. & Top-5 Acc.\\
      \midrule
      {\bfseries Cat2Real-DINOv3-384} & {\bfseries 80.73} & {\bfseries 94.70}\\
      Stage 1 checkpoint & 73.77 & 93.24\\
      Stage 2 checkpoint & 78.45 & 93.65\\
      No similar product selection & 74.81 & 93.41\\
      No closest image selection & 77.42 & 93.52\\
      No dynamic closest image & 78.65 & 93.91\\
      No low-quality image filter & 79.52 & 94.21\\
      1 random negative image & 73.78 & 90.15\\
      4 selected negative images & 76.37 & 94.06\\
      \bottomrule
    \end{tabular}
    \caption{\textbf{Ablation study of the Cat2Real training framework.} Performance impact when isolating individual iterative training stages and different training configurations.}
    \label{tab:ablation}
\end{table}

\begin{figure}[ht]
  \centering
    \includegraphics[width=0.9\columnwidth]{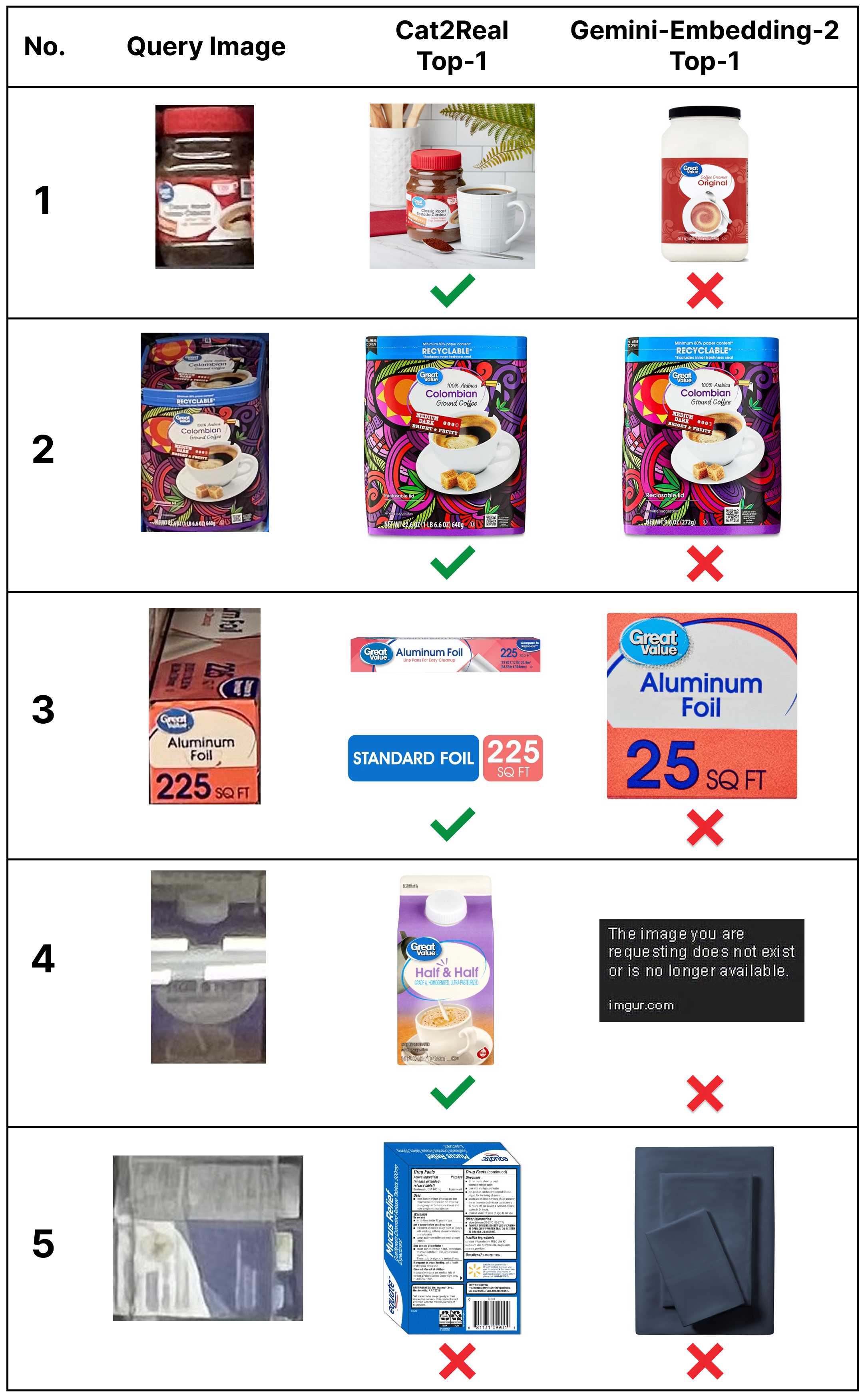}
    \caption{
      \textbf{Qualitative comparison of Top-1 product recognition predictions.} Visual examples highlight predictions from the proposed Cat2Real model against the proprietary Gemini-Embedding-2 baseline.
    }
    \label{fig:4}
\end{figure}

The comparative evaluation on our internal product recognition evaluation dataset is summarized in \cref{tab:1}, demonstrating the substantial efficacy of the proposed Cat2Real three-stage contrastive learning framework.
 
Out-of-the-box foundation models, such as DINOv2 and DINOv3, perform poorly on this specialized retail classification task due to the absence of domain-specific optimization. For instance, the DINOv3 with an image size of 384 baseline yields a modest Top-1 accuracy of only 53.83\%. In contrast, adapting these models via the Cat2Real paradigm yields massive performance gains across all architectures and resolutions. The Cat2Real-DINOv3-384 configuration delivers the highest performance, securing an 80.73\% Top-1 accuracy and a 94.70\% Top-5 accuracy. Notably, all fine-tuned Cat2Real models significantly outperform state-of-the-art commercial embedding models, including Gemini-Embedding-2 (68.71\% Top-1) and Multimodalembedding@001 (63.51\% Top-1). Detailed per-category evaluation metrics comparing Cat2Real-DINOv3-384 and Gemini-Embedding-2 are provided in \cref{tab:eval_metrics} in the Supplementary Material.

The robust improvements across all configurations indicate that the three-stage contrastive learning strategy successfully reconciles the distribution gap between high-quality catalog packshots and noisy, real-world product crops. By exploiting product- and image-level similarities for hard negative mining, the framework enables the encoder to extract highly discriminative features, resolving fine-grained ambiguities under challenging in-store noise (e.g., severe blur, extreme rotations, and partial occlusions).

The qualitative advantages of the Cat2Real framework are illustrated via the visual examples in \cref{fig:4}, which contrasts the Top-1 predictions of our model against the Gemini-Embedding-2 baseline:
 
\textbf{Fine-Grained Discrimination:} As shown in rows 1--3, the Cat2Real model accurately identifies the target products across diverse camera perspectives. Conversely, the commercial baseline matches items based on coarse visual similarity, leading to incorrect product IDs. These cases demonstrate our model's capacity to recognize subtle patterns, such as size variations or brand sub-line text, which are typically obscured in real-world crops. This confirms that Cat2Real successfully learns which semantic attributes to prioritize (e.g., distinguishing product volume over viewpoint changes)—a capacity lacking in generic embedding models.

\textbf{Robustness to Environmental Noise:} Row 4 underscores the model's resilience to extreme acquisition noise. Despite the query image being heavily blurred, Cat2Real successfully retrieves the exact match, whereas the baseline returns an entirely unrelated product class. Row 5 details a common failure mode where extreme blur eliminates all distinguishing features. While both models fail, the Cat2Real prediction remains bounded within a closely related visual and functional domain, whereas the baseline degrades completely to irrelevant categories (e.g., linens).

These qualitative results corroborate that aligning disparate data distributions into a unified latent space equips Cat2Real with superior discriminatory power for large-scale retail recognition.

\subsection{Generalization and Scaling Performance}
To evaluate the generalization and scalability potential of Cat2Real for large-scale retail ecosystems, we analyze performance across varying training dataset scales and compositions, as detailed in \cref{tab:scaling}. The results demonstrate the framework's capability to scale up and seamlessly generalize to entirely unseen products and categories without requiring model parameter updates.

\textbf{Generalization to Unseen Products and Categories:} Continuous inventory turnover and frequent product launches render frequent model retraining impractical for global retailers. To assess the zero-shot adaptability of our framework, we evaluate a ``Remove Eval Products'' protocol. In this setup, all products present in the evaluation set are completely purged from the training phase, reducing the number of unique training items from 45,159 to 43,330. Despite having zero exposure to these specific products or their corresponding real-world crops during training, the model exhibits remarkable resilience, experiencing only a negligible performance drop. It delivers a high Top-1 accuracy of 80.23\% (compared to 80.73\% for the full training setup) and a Top-5 accuracy of 94.20\%. Furthermore, zero-shot generalization across distinct merchandise categories represents an even steeper challenge due to substantial domain shifts in packaging and geometries. Notably, models fine-tuned on subsets (Batches 1--4) show highly favorable zero-shot transferability to completely unencountered categories. This indicates that given a sufficiently diverse and multi-category training distribution, the Cat2Real objective successfully extracts fundamental, domain-invariant features for fine-grained discrimination that generalize robustly to novel product categories.

\textbf{Performance with Data Volume Scaling:} We observe a monotonically increasing trend in recognition accuracy as the volume and categorical diversity of the training data scale up. When trained on the restricted ``Batch 1'' subset—comprising only 6 categories and 11,206 products, the model establishes a baseline Top-1 accuracy of 75.77\%. Incrementally expanding the dataset breadth through Batches 2, 3, and 4 drives a steady climb in performance, reaching 78.89\% Top-1 accuracy at 31 categories. Peak performance is unlocked when training on the full corpus (52 categories, 45,243 unique products, and 196,531 real-world images), culminating in an 80.73\% Top-1 and 94.70\% Top-5 accuracy. This evidence shows that the Cat2Real paradigm has the potential for continuous improvement as more data from different products and categories is collected.

These findings validate that the proposed three-stage contrastive learning paradigm successfully addresses long-standing scalability hurdles. By supervising the network to project real-world product crops and multi-view catalog packshots into a unified latent space, Cat2Real handles open-world inventory expansion efficiently and economically, making it uniquely suited for continuous commercial deployment.

\subsection{Ablation Studies}
To systematically evaluate the contribution of each individual component within the Cat2Real framework, we perform an ablation analysis using the DINOv3 architecture. The results, summarized in \cref{tab:ablation}, substantiate that our multi-stage training progression and specialized sampling strategies are indispensable for achieving top-tier accuracy in production-level retail environments.

\subsubsection{Impact of the Multi-Stage Framework}
The progressive training paradigm demonstrates a clear trajectory of performance gains as the sampling criteria advance from naive selection to dynamic, model-driven mining. The Stage 1 checkpoint establishes a baseline Top-1 accuracy of 73.77\% by leveraging only primary catalog images and category-level negative pairs. Transitioning to Stage 2 introduces image-level similarity filtering and targets hard negative items from visually similar products, which significantly lifts the Top-1 accuracy to 78.45\%. Finally, the full Cat2Real model reaches its peak performance of 80.73\% in Stage 3 by executing dynamic image selection, which continuously refines the latent space by re-evaluating and selecting the optimal positive and negative pairs on the fly using the updated encoder weights.

\subsubsection{Impact of Hard Negative Mining}
Our strategic image sampling scheme serves as the core driver for fine-grained discrimination. Eliminating similar-product sampling leads to a substantial performance drop, with Top-1 accuracy falling to 74.81\%, which demonstrates that the model requires explicit exposure to near-identical items to map fine-grained boundary features. Performance also declines to 77.42\% when the viewpoint-matching closest image selection is removed, and to 78.65\% if this selection is frozen statically rather than updated dynamically. These strategies ensure that both positive and negative catalog views are tightly aligned with the query perspective, forcing the encoder to concentrate on invariant semantic details rather than geometric or perspective disparities.

\subsubsection{Impact of Low-Quality Image Filter}
Filtering out unmatchable or corrupted query crops is essential for stabilizing the optimization trajectory and minimizing gradient noise. Omitting the low-quality image removal step causes the Top-1 accuracy to decrease to 79.52\%. This filtering mechanism identifies real-world crops whose maximum similarity score to any catalog reference falls below the pre-established threshold $T_I$, flagging instances of severe motion blur, occlusion, or mislabeling. Purging these uninformative samples prevents the network from executing erratic optimization steps driven by corrupted gradients.

\subsubsection{Impact of the Loss Formulation}
The targeted triplet margin loss demonstrates clear advantages over alternative contrastive loss structures that rely on wider or randomized negative sets. Relying on standard random negative selection—a common convention in generic contrastive learning—yields a significantly lower Top-1 accuracy of 73.78\%. Furthermore, extending the formulation to contrast the query against four negative products simultaneously rather than a single one drops the accuracy to 76.37\%. These results confirm that a focused, triplet-based margin objective that isolates and pushes the hard negative sample away from the query-positive anchor is optimal for this task.
\section{Conclusions}
\label{sec:conclusions}

In summary, Cat2Real presents a novel and scalable contrastive learning framework that addresses the significant challenges of large-scale product recognition in retail by redefining the task as an embedding-based similarity search between noisy real-world images and high-resolution catalog packshots. This approach eliminates the need for continuous model retraining as inventories change, utilizing a three-stage training paradigm to iteratively fine-tune image encoders such as DINOv2 and DINOv3. By leveraging product-level and image-level similarities for intentional data sampling and employing a dynamic image selection strategy, Cat2Real achieves a Top-1 accuracy of 80.73\% and a Top-5 accuracy of 94.70\%, significantly outperforming state-of-the-art foundation models. Furthermore, our evaluation demonstrates that the model generalizes effectively to unseen products and categories with minimal performance loss, providing a robust and economical solution for continuous, large-scale retail deployments.
{
    \small
    \bibliographystyle{ieeenat_fullname}
    \bibliography{main}

@String(BMVC= {Brit. Mach. Vis. Conf.})

@String(BMVC  =	{BMVC})

@INPROCEEDINGS{10533619,
  author={Varghese, Rejin and M., Sambath},
  booktitle={2024 International Conference on Advances in Data Engineering and Intelligent Computing Systems (ADICS)}, 
  title={YOLOv8: A Novel Object Detection Algorithm with Enhanced Performance and Robustness}, 
  year={2024},
  volume={},
  number={},
  pages={1-6},
  doi={10.1109/ADICS58448.2024.10533619}
}

@ARTICLE{10528273,
  author={Alghamdi, Mohammed and Mengash, Hanan Abdullah and Aljebreen, Mohammed and Maray, Mohammed and Darem, Abdulbasit A. and Salama, Ahmed S.},
  journal={IEEE Access}, 
  title={Empowering Retail Through Advanced Consumer Product Recognition Using Aquila Optimization Algorithm With Deep Learning}, 
  year={2024},
  volume={12},
  number={},
  pages={71055-71065},
  doi={10.1109/ACCESS.2024.3399480}
}

@inproceedings{NIPS2015_14bfa6bb,
 author = {Ren, Shaoqing and He, Kaiming and Girshick, Ross and Sun, Jian},
 booktitle = {Advances in Neural Information Processing Systems},
 editor = {C. Cortes and N. Lawrence and D. Lee and M. Sugiyama and R. Garnett},
 pages = {},
 publisher = {Curran Associates, Inc.},
 title = {Faster R-CNN: Towards Real-Time Object Detection with Region Proposal Networks},
 url = {https://proceedings.neurips.cc/paper_files/paper/2015/file/14bfa6bb14875e45bba028a21ed38046-Paper.pdf},
 volume = {28},
 year = {2015}
}

@misc{sinha2022improveddeeplearningapproach,
      title={An Improved Deep Learning Approach For Product Recognition on Racks in Retail Stores}, 
      author={Ankit Sinha and Soham Banerjee and Pratik Chattopadhyay},
      year={2022},
      eprint={2202.13081},
      archivePrefix={arXiv},
      primaryClass={cs.CV},
      url={https://arxiv.org/abs/2202.13081}
}

@ARTICLE{FAM1,
  author={Prabu Selvam and Joseph Abraham Sundar Koilraj},
  journal={Food Analytical Methods}, 
  title={A Deep Learning Framework for Grocery Product Detection and Recognition}, 
  year={2022},
  volume={15},
  number={},
  pages={3498–3522},
  doi={10.1007/s12161-022-02384-2}
}

@misc{khanam2024yolov5deeplookinternal,
      title={What is YOLOv5: A deep look into the internal features of the popular object detector}, 
      author={Rahima Khanam and Muhammad Hussain},
      year={2024},
      eprint={2407.20892},
      archivePrefix={arXiv},
      primaryClass={cs.CV},
      url={https://arxiv.org/abs/2407.20892}, 
}

@ARTICLE{MVA1,
  author={Tobias Pettersson and Maria Riveiro and Tuwe Löfström},
  journal={Machine Vision and Applications}, 
  title={Multimodal fine-grained grocery product recognition using image and OCR text}, 
  year={2024},
  volume={35},
  number={79},
  doi={10.1007/s00138-024-01549-9}
}

@article{OUCHEIKH2022115942,
title = {Product verification using OCR classification and Mondrian conformal prediction},
journal = {Expert Systems with Applications},
volume = {188},
pages = {115942},
year = {2022},
issn = {0957-4174},
doi = {https://doi.org/10.1016/j.eswa.2021.115942},
url = {https://www.sciencedirect.com/science/article/pii/S0957417421012963},
author = {Rachid Oucheikh and Tobias Pettersson and Tuwe Löfström}
}

@article{vit,
  author       = {Alexey Dosovitskiy and
                  Lucas Beyer and
                  Alexander Kolesnikov and
                  Dirk Weissenborn and
                  Xiaohua Zhai and
                  Thomas Unterthiner and
                  Mostafa Dehghani and
                  Matthias Minderer and
                  Georg Heigold and
                  Sylvain Gelly and
                  Jakob Uszkoreit and
                  Neil Houlsby},
  title        = {An Image is Worth 16x16 Words: Transformers for Image Recognition
                  at Scale},
  journal      = {CoRR},
  volume       = {abs/2010.11929},
  year         = {2020},
  url          = {https://arxiv.org/abs/2010.11929},
  eprinttype   = {arXiv},
  eprint       = {2010.11929},
  bibsource    = {dblp computer science bibliography, https://dblp.org}
}

@article{clip,
  author       = {Alec Radford and
                  Jong Wook Kim and
                  Chris Hallacy and
                  Aditya Ramesh and
                  Gabriel Goh and
                  Sandhini Agarwal and
                  Girish Sastry and
                  Amanda Askell and
                  Pamela Mishkin and
                  Jack Clark and
                  Gretchen Krueger and
                  Ilya Sutskever},
  title        = {Learning Transferable Visual Models From Natural Language Supervision},
  journal      = {CoRR},
  volume       = {abs/2103.00020},
  year         = {2021},
  url          = {https://arxiv.org/abs/2103.00020},
  eprinttype   = {arXiv},
  eprint       = {2103.00020},
  bibsource    = {dblp computer science bibliography, https://dblp.org}
}

@article{mae,
  author       = {Kaiming He and
                  Xinlei Chen and
                  Saining Xie and
                  Yanghao Li and
                  Piotr Doll{\'{a}}r and
                  Ross B. Girshick},
  title        = {Masked Autoencoders Are Scalable Vision Learners},
  journal      = {CoRR},
  volume       = {abs/2111.06377},
  year         = {2021},
  url          = {https://arxiv.org/abs/2111.06377},
  eprinttype   = {arXiv},
  eprint       = {2111.06377},
  bibsource    = {dblp computer science bibliography, https://dblp.org}
}

@article{dino,
  author       = {Mathilde Caron and
                  Hugo Touvron and
                  Ishan Misra and
                  Herv{\'{e}} J{\'{e}}gou and
                  Julien Mairal and
                  Piotr Bojanowski and
                  Armand Joulin},
  title        = {Emerging Properties in Self-Supervised Vision Transformers},
  journal      = {CoRR},
  volume       = {abs/2104.14294},
  year         = {2021},
  url          = {https://arxiv.org/abs/2104.14294},
  eprinttype   = {arXiv},
  eprint       = {2104.14294},
  bibsource    = {dblp computer science bibliography, https://dblp.org}
}

@misc{dinov2,
      title={DINOv2: Learning Robust Visual Features without Supervision}, 
      author={Maxime Oquab and Timothée Darcet and Théo Moutakanni and Huy Vo and Marc Szafraniec and Vasil Khalidov and Pierre Fernandez and Daniel Haziza and Francisco Massa and Alaaeldin El-Nouby and Mahmoud Assran and Nicolas Ballas and Wojciech Galuba and Russell Howes and Po-Yao Huang and Shang-Wen Li and Ishan Misra and Michael Rabbat and Vasu Sharma and Gabriel Synnaeve and Hu Xu and Hervé Jegou and Julien Mairal and Patrick Labatut and Armand Joulin and Piotr Bojanowski},
      year={2024},
      eprint={2304.07193},
      archivePrefix={arXiv},
      primaryClass={cs.CV},
      url={https://arxiv.org/abs/2304.07193}, 
}

@misc{dinov3,
      title={DINOv3}, 
      author={Oriane Siméoni and Huy V. Vo and Maximilian Seitzer and Federico Baldassarre and Maxime Oquab and Cijo Jose and Vasil Khalidov and Marc Szafraniec and Seungeun Yi and Michaël Ramamonjisoa and Francisco Massa and Daniel Haziza and Luca Wehrstedt and Jianyuan Wang and Timothée Darcet and Théo Moutakanni and Leonel Sentana and Claire Roberts and Andrea Vedaldi and Jamie Tolan and John Brandt and Camille Couprie and Julien Mairal and Hervé Jégou and Patrick Labatut and Piotr Bojanowski},
      year={2025},
      eprint={2508.10104},
      archivePrefix={arXiv},
      primaryClass={cs.CV},
      url={https://arxiv.org/abs/2508.10104}, 
}

@article{SimCLR,
  author       = {Ting Chen and
                  Simon Kornblith and
                  Mohammad Norouzi and
                  Geoffrey E. Hinton},
  title        = {A Simple Framework for Contrastive Learning of Visual Representations},
  journal      = {CoRR},
  volume       = {abs/2002.05709},
  year         = {2020},
  url          = {https://arxiv.org/abs/2002.05709},
  eprinttype   = {arXiv},
  eprint       = {2002.05709},
  bibsource    = {dblp computer science bibliography, https://dblp.org}
}

@article{moco,
  author       = {Kaiming He and
                  Haoqi Fan and
                  Yuxin Wu and
                  Saining Xie and
                  Ross B. Girshick},
  title        = {Momentum Contrast for Unsupervised Visual Representation Learning},
  journal      = {CoRR},
  volume       = {abs/1911.05722},
  year         = {2019},
  url          = {http://arxiv.org/abs/1911.05722},
  eprinttype   = {arXiv},
  eprint       = {1911.05722},
  bibsource    = {dblp computer science bibliography, https://dblp.org}
}

@article{infonce,
  author       = {A{\"{a}}ron van den Oord and
                  Yazhe Li and
                  Oriol Vinyals},
  title        = {Representation Learning with Contrastive Predictive Coding},
  journal      = {CoRR},
  volume       = {abs/1807.03748},
  year         = {2018},
  url          = {http://arxiv.org/abs/1807.03748},
  eprinttype   = {arXiv},
  eprint       = {1807.03748},
  bibsource    = {dblp computer science bibliography, https://dblp.org}
}

@inproceedings{nt_xent,
 author = {Sohn, Kihyuk},
 booktitle = {Advances in Neural Information Processing Systems},
 editor = {D. Lee and M. Sugiyama and U. Luxburg and I. Guyon and R. Garnett},
 pages = {},
 publisher = {Curran Associates, Inc.},
 title = {Improved Deep Metric Learning with Multi-class N-pair Loss Objective},
 url = {https://proceedings.neurips.cc/paper_files/paper/2016/file/6b180037abbebea991d8b1232f8a8ca9-Paper.pdf},
 volume = {29},
 year = {2016}
}

@inproceedings{cnn,
 author = {Krizhevsky, Alex and Sutskever, Ilya and Hinton, Geoffrey E},
 booktitle = {Advances in Neural Information Processing Systems},
 editor = {F. Pereira and C.J. Burges and L. Bottou and K. Weinberger},
 pages = {},
 publisher = {Curran Associates, Inc.},
 title = {ImageNet Classification with Deep Convolutional Neural Networks},
 url = {https://proceedings.neurips.cc/paper_files/paper/2012/file/c399862d3b9d6b76c8436e924a68c45b-Paper.pdf},
 volume = {25},
 year = {2012}
}

@article{resnet,
  author       = {Kaiming He and
                  Xiangyu Zhang and
                  Shaoqing Ren and
                  Jian Sun},
  title        = {Deep Residual Learning for Image Recognition},
  journal      = {CoRR},
  volume       = {abs/1512.03385},
  year         = {2015},
  url          = {http://arxiv.org/abs/1512.03385},
  eprinttype   = {arXiv},
  eprint       = {1512.03385},
  bibsource    = {dblp computer science bibliography, https://dblp.org}
}

@article{facenet,
  author       = {Florian Schroff and
                  Dmitry Kalenichenko and
                  James Philbin},
  title        = {FaceNet: {A} Unified Embedding for Face Recognition and Clustering},
  journal      = {CoRR},
  volume       = {abs/1503.03832},
  year         = {2015},
  url          = {http://arxiv.org/abs/1503.03832},
  eprinttype   = {arXiv},
  eprint       = {1503.03832},
  bibsource    = {dblp computer science bibliography, https://dblp.org}
}

@inproceedings{triplet,
        	title={Learning local feature descriptors with triplets and shallow convolutional neural networks},
        	author={Vassileios Balntas, Edgar Riba, Daniel Ponsa and Krystian  Mikolajczyk},
        	year={2016},
        	month={September},
        	pages={119.1-119.11},
        	articleno={119},
        	numpages={11},
        	booktitle={Proceedings of the British Machine Vision Conference (BMVC)},
        	publisher={BMVA Press},
        	editor={Richard C. Wilson, Edwin R. Hancock and William A. P. Smith},
        	doi={10.5244/C.30.119},
        	isbn={1-901725-59-6},
        	url={https://dx.doi.org/10.5244/C.30.119}
        }

@article{metalearning,
  author       = {Yinbo Chen and
                  Xiaolong Wang and
                  Zhuang Liu and
                  Huijuan Xu and
                  Trevor Darrell},
  title        = {A New Meta-Baseline for Few-Shot Learning},
  journal      = {CoRR},
  volume       = {abs/2003.04390},
  year         = {2020},
  url          = {https://arxiv.org/abs/2003.04390},
  eprinttype   = {arXiv},
  eprint       = {2003.04390},
  bibsource    = {dblp computer science bibliography, https://dblp.org}
}

@misc{geminiembedding2,
      title={Gemini Embedding 2: A Native Multimodal Embedding Model from Gemini}, 
      author={Madhuri Shanbhogue and Zhe Li and Shanfeng Zhang and Gustavo Hernández Ábrego and Shih-Cheng Huang and Aashi Jain and Daniel Salz and Sonam Goenka and Chaitra Hegde and Ji Ma and Feiyang Chen and Jiaxing Wu and Tanmaya Dabral and Babak Samari and Kevin Poulet and Daniel Cer and Kaifeng Chen and Paul Suganathan and Hui Hui and Jovan Andonov and Philippe Schlattner and Jay Han and Iftekhar Naim and Wing Lowe and Vladimir Pchelin and Albert Yang and Yi-Ting Chen and Zhongli Ding and Grace Zhang and Georg Heigold and Yichang Chen and Antoine Reveillon and Brendan Mccloskey and Wenlei Zhou and Dahun Kim and Rui Meng and Emma Wang and Jack Zheng and Halley Fede and Zhen Yang and Keegan Mosley and Brian Potetz and Sahil Dua and Henrique Schechter Vera and Shen Gao and Hesen Zhang and Andreas Hess and Hengxuan Ying and Alberto Montes and Karan Gill and Min Choi and Sebastian Russo and Anja Hauth and Jinhyuk Lee and Michael Boratko and Megan Barnes and Vikram Rao and Claudiu Musat and Cyril Allauzen and Ehsan Variani and Shankar Kumar and Tom Bagby and Junyi Jiao and Yang Gu and Tengxin Li and Ayush Agrawal and Roberto Santana and Dev Nath and Stephen Karukas and Shuoxuan Han and Lucia Loher and Alice Twu and Nidhi Vyas and Siddharth Bhai and Frank Palma Gomez and Wangyuan Zhang and Chaoren Liu and Jizheng Yang and Steve Qiu and Shijie Zhang and Sujay Kulkarni and Sascha Rothe and Sean Nakamoto and Raphael Hoffmann and Zach Gleicher and Yunhsuan Sung and Qin Yin and Tom Duerig and Mojtaba Seyedhosseini},
      year={2026},
      eprint={2605.27295},
      archivePrefix={arXiv},
      primaryClass={cs.CV},
      url={https://arxiv.org/abs/2605.27295}, 
}
}

\newpage
\appendix
\onecolumn
\section{Supplementary Material}


\begin{table}[H]
  \centering 
    \begin{tabular}{@{}lcr@{}}
      \toprule
       & Training Set & Evaluation Set\\
      \midrule
      No. of Real-World Images & 196,531 & 5,043\\
      No. of Unique Products with Real-World Images  & 26,910 & 1,865\\
      No. of Catalog Images & 409,891 & 18,393\\
      No. of Unique Products with Catalog Images & 45,243 & 2,155\\
      \bottomrule
    \end{tabular}
    \caption{\textbf{Quantitative breakdown and statistics of the product recognition dataset.} Detailed counts for the real-world query crops, reference catalog packshots, and their unique product IDs across both training and evaluation subsets.}
    \label{tab:dataset_stat}
\end{table}

\begin{table*}[t]
  \centering
  \small
  \begin{tabular}{@{}lccccc@{}}
    \toprule
    & & \multicolumn{2}{c}{Cat2Real-DINOv3-384} & \multicolumn{2}{c}{Gemini-Embedding-2} \\
    \cmidrule(lr){3-4} \cmidrule(lr){5-6}
    Category & Image Count & Top-1 Acc. & Top-5 Acc. & Top-1 Acc. & Top-5 Acc.\\
    \midrule
    Additives & 24 & 91.67 & 100.00 & 79.17 & 100.00\\
    Air Fresheners And Deodorizers & 29 & 96.55 & 100.00 & 96.55 & 96.55\\
    All Purpose Cleaners & 183 & 91.26 & 100.00 & 78.69 & 98.36\\
    Allergy & 200 & 82.50 & 93.50 & 51.00 & 80.00\\
    Analgesics & 164 & 60.98 & 88.41 & 15.24 & 46.34\\
    Auto Motor Oil And Lubricants & 200 & 51.50 & 86.50 & 52.50 & 83.50\\
    Bed Pillows & 184 & 76.09 & 94.02 & 65.22 & 96.20\\
    Bed Sheets & 200 & 60.50 & 78.50 & 55.50 & 75.00\\
    Body Wash & 61 & 65.57 & 78.69 & 37.70 & 54.10\\
    Cheeses & 200 & 61.50 & 90.50 & 52.00 & 84.50\\
    Coffee & 200 & 82.00 & 95.00 & 63.00 & 84.00\\
    Cold Cereal & 63 & 93.65 & 100.00 & 84.13 & 100.00\\
    Cookies & 153 & 91.50 & 98.69 & 90.20 & 97.39\\
    Cough Cold Flu & 200 & 69.00 & 90.00 & 35.50 & 64.50\\
    Crackers & 56 & 85.71 & 92.86 & 80.36 & 89.29\\
    Dish Care & 30 & 93.33 & 100.00 & 90.00 & 100.00\\
    Facecare & 55 & 83.64 & 100.00 & 81.82 & 98.18\\
    Foils Bags And Wraps & 200 & 64.50 & 82.50 & 50.00 & 72.00\\
    Food Preservation & 158 & 66.46 & 81.65 & 61.39 & 82.91\\
    Frozen Breakfast & 83 & 90.36 & 97.59 & 91.57 & 98.80\\
    Frozen Core Meals & 103 & 98.06 & 99.03 & 91.26 & 97.09\\
    Frozen Pizza & 94 & 78.72 & 95.74 & 94.68 & 98.94\\
    Frozen Snacks & 83 & 89.16 & 98.80 & 78.31 & 90.36\\
    Hand And Body Lotion & 38 & 89.47 & 97.37 & 73.68 & 100.00\\
    Heated Cooking & 32 & 81.25 & 100.00 & 78.12 & 100.00\\
    Lifestyle Nutrition & 34 & 55.88 & 91.18 & 52.94 & 94.12\\
    Lunch Meat & 98 & 84.69 & 93.88 & 66.33 & 90.82\\
    Mainstream Chilled & 40 & 95.00 & 100.00 & 92.50 & 97.50\\
    Milk Creamers & 200 & 68.00 & 95.50 & 51.50 & 89.00\\
    Milk Specialty & 151 & 94.04 & 100.00 & 84.77 & 100.00\\
    Mouthwash And Breath & 51 & 92.16 & 100.00 & 84.31 & 96.08\\
    Multi Salty & 140 & 78.57 & 97.14 & 72.14 & 95.00\\
    Oil And Shortening & 129 & 86.82 & 100.00 & 71.32 & 96.90\\
    Period Care & 129 & 75.19 & 95.35 & 64.34 & 89.92\\
    Salty Snacks & 116 & 87.07 & 99.14 & 75.86 & 95.69\\
    Soup & 162 & 83.33 & 96.30 & 67.28 & 93.21\\
    Sweet Baked Goods & 37 & 97.30 & 97.30 & 72.97 & 89.19\\
    Tacos Enchiladas And Salsa & 98 & 85.71 & 94.90 & 61.22 & 81.63\\
    Toast And Bread & 24 & 75.00 & 91.67 & 70.83 & 91.67\\
    Toothpaste & 88 & 78.41 & 94.32 & 78.41 & 100.00\\
    Trash Bags & 162 & 82.10 & 99.38 & 49.38 & 96.30\\
    Vitamins And Supplements & 200 & 68.50 & 90.00 & 34.00 & 64.50\\
    Yogurt & 191 & 88.48 & 96.86 & 76.96 & 92.15\\
    \midrule
    \textbf{Macro Average} &  & \textbf{80.73} & \textbf{94.70} & \textbf{68.71} & \textbf{89.34}\\
    \bottomrule
  \end{tabular}
  \caption{\textbf{Detailed per-category fine-grained evaluation metrics.} A comprehensive comparison of top-1 and top-5 accuracies across individual retail inventory categories between the Cat2Real-DINOv3-384 model and the Gemini-Embedding-2 baseline.}
  \label{tab:eval_metrics}
\end{table*}

\end{document}